\documentclass[preprint,12pt]{elsarticle}

\usepackage{auto-pst-pdf}

\usepackage{url}
\usepackage{graphicx}
\usepackage{epsfig}
\usepackage{url}
\usepackage{textcomp}
\usepackage{booktabs}
\usepackage{pstricks}
\usepackage{setspace}
\usepackage{natbib}
\usepackage{color}
\usepackage{amsfonts}
\usepackage{amsmath,amsthm,amscd}

\newcommand{\pa}{\partial}

\newtheorem{theorem}{Theorem}

\newtheorem{definition}{Definition}
\newtheorem{lemma}{Lemma}
\newtheorem{remark}{Remark}

\newcommand{\be}{\begin{equation}}
\newcommand{\ee}{\end{equation}}

\journal{Neural Networks}

\begin{document}

\begin{frontmatter}



\title{Complex-Valued Autoencoders}


\author{Pierre Baldi$^a$}
\author{Zhiqin Lu$^b$}
\address[l1]{Department of Computer Science, UCI, Irvine, CA 92697-3435}
\address[l2]{Department of Mathematics, UCI, Irvine, CA 92697-3875}

\begin{abstract}
Autoencoders are unsupervised machine learning circuits, with typically one hidden layer, whose learning goal is to minimize an average distortion measure between inputs and outputs. Linear autoencoders correspond to the special case where only linear transformations between visible and hidden variables are used. While linear autoencoders can be defined over any field, only real-valued linear autoencoders have been studied so far. Here we study complex-valued
linear autoencoders where the components of the training vectors and adjustable matrices are defined over the complex field with the $L_2$ norm. We provide simpler and more general proofs that unify the real-valued and complex-valued cases, showing that in both cases the landscape of the error function is invariant under certain groups of transformations. The landscape has no local minima, a family of global minima associated with Principal Component Analysis, and many families of saddle points associated with orthogonal projections onto sub-space spanned by sub-optimal subsets of eigenvectors of the covariance matrix. The theory yields several iterative, convergent, learning algorithms, a clear understanding of the generalization properties of the trained autoencoders, and can equally be applied to the hetero-associative case when external targets are provided. Partial results on deep architecture as well as the differential geometry of autoencoders are also presented. The general framework described here is useful to classify autoencoders and identify general properties that ought to be investigated for each class, illuminating some of the connections between autoencoders, unsupervised learning, clustering, Hebbian learning, and information theory.
\end{abstract}

\end{frontmatter}


\section{Introduction}
\label{intro}

Autoencoder circuits, which try to minimize a distortion measure between inputs and outputs, play a fundamental role in machine learning. They were introduced in the 1980s by the Parallel Distributed Processing (PDP) group \cite{rumelhart1986learning}  as a way to address the problem of unsupervised learning, in contrast to supervised learning in backpropagation networks, by using the inputs as learning targets.
More recently, autoencoders have been used extensively in the ``deep architecture'' approach
\cite{hinton2006fast,hinton2006reducing,bengio-lecun-07,Erhan+al-2010}, where autoencoders in the form of Restricted Boltzmann Machines (RBMS) are stacked
and trained bottom up in unsupervised fashion to extract hidden features and efficient representations
that can then be used to address supervised classification or regression tasks.
In spite of the interest they have generated, and with a few
exceptions \cite{bengio2010}, little theoretical understanding of autoencoders and deep architectures has been obtained to date.
One possible strategy for addressing these issues is to partition the autoencoder universe into different classes, for instance linear versus non-linear autoencoders, and identify classes that can be analyzed mathematically, with the hope that the
precise understanding of several specific classes may lead to a clearer general picture.
Within this background and strategy, the main purpose of this article is to provide a complete theory for a particular class of autoencoders, namely linear autoencoders over the complex field.

In addition to trying to progressively derive a more complete theoretical understanding of autoencoders, there are several other reasons, primarily theoretical ones,  for looking at linear complex-valued autoencoders. First, linear autoencoders over the real numbers were solved by Baldi and Hornik \cite{baldihornik88} (see also \cite{bourlard1988auto}). It is thus natural to ask whether linear autoencoders over the complex numbers share the same basic properties or not, and whether unified proofs can be derived to cover both the real- and complex-valued cases. More generally linear autoencoders can be defined over any field and therefore one can raise similar questions for linear autoencoders over other fields, such as finite Galois fields \cite{lang1984algebra}.

Second, a specific class of non-linear autoencoders was recently introduced and analyzed mathematically \cite{baldijmlr11}. This is the class of Boolean autoencoders where all circuit operations are Boolean functions. It can be shown that this class of autoencoders is intimately connected to clustering and so it is reasonable to both compare Boolean autoencoders to linear autoencoders, and to examine linear autoencoders from a clustering perspective.

Third, there has been a trend in recent years towards the use of linear networks and methods to address difficult tasks, such as building recommender systems (e.g. the Netflix prize challenge \cite{bell2007lessons,takacs2008matrix}) or modeling the development of sensory systems, in clever ways by introducing particular restrictions on the relevant matrices, such as sparsity or low-rank \cite{candes2008people,candes2009exact}. Autoencoders discussed in this paper can be viewed as linear, low-rank, approximations to the identity function and therefore fall within this general trend.

Finally, complex vector spaces and matrices have several areas of specific application, ranging from quantum mechanics, to fast Fourier transforms, to complex-valued neural networks \cite{hirose2003complex}, and ought to be studied in their own right. Complex-valued linear autoencoders can be viewed as a particular class of complex-valued neural networks and may be used
in applications involving complex-valued data.

With these motivations in mind, in order to provide a complete treatment of linear complex-valued autoencoders here
we first introduce a general framework and notation, essential for a better understanding and classification of autoencoders, and for the identification of common properties that ought to be studied in any new specific autoencoder case. We then proceed to analytically solve the complex-valued linear autoencoder.
While in the end the results obtained in the complex-valued case are similar to those previously obtained in the real-valued case \cite{baldihornik88} interchanging conjugate transposition with simple transposition, the approach adopted here allow us to derive simpler and more general proofs that unify both cases. In addition, we derive several new properties and results, addressing for instance learning algorithms and their convergence properties, and some of the connections to clustering, deep architectures, and other kinds of autoencoders. Finally, in the Appendix, we begin the study of real- and complex-valued autoencoders from a differential geometry perspective.

\section{General Autoencoder Framework and Preliminaries}

\subsection{General Autoencoder Framework}
\begin{figure}[ht]
\begin{center}
\includegraphics[width=0.50\columnwidth]{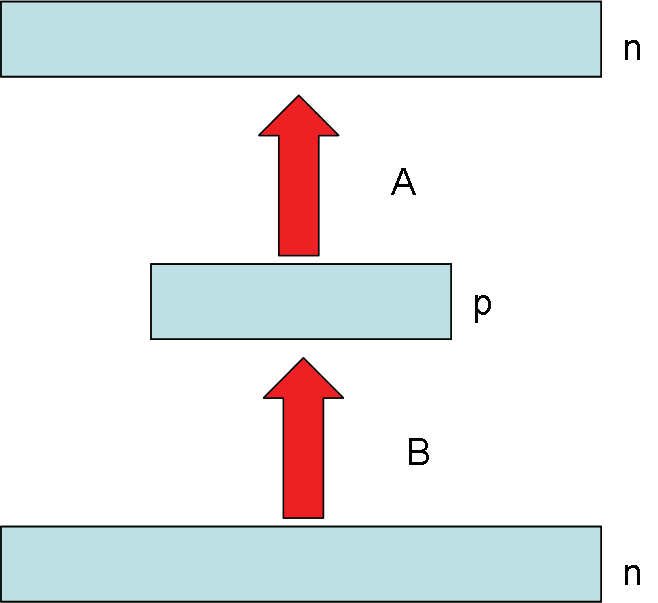}
\end{center}
\caption{An $n/p/n$ Autoencoder Architecture.}
\label{fig:autoencoder}
\end{figure}

To derive a fairly general framework, an $n/p/n$ {autoencoder} (Figure \ref{fig:autoencoder}) is defined by a t-uple ${\mathbb F}, {\mathbb G}, n, p, {\cal A}, {\cal B}, {\cal X}, \Delta$ where:
\begin{enumerate}
\item $\mathbb F$ and $\mathbb G$ are sets.
\item $n$ and $p$ are positive integers. Here we consider primarily the case where $0<p<n$.
\item $\cal A$ is a class of functions from ${\mathbb G}^p$ to ${\mathbb F}^n$.
\item $\cal B$ is a class of functions from ${\mathbb F}^n$ to ${\mathbb G}^p$.
\item ${\cal X}=\{{ x}_1,\ldots,{ x}_m\}$ is a set of $m$
(training) vectors in ${ \mathbb F}^n$. When external targets are present, we let
 ${\cal Y}=\{{y}_1,\ldots,{\ y}_m\}$ denote the corresponding set of target vectors in ${\mathbb F}^n$.
\item $\Delta$ is a dissimilarity or distortion function defined over ${\mathbb F}^n$.
\end{enumerate}
For any ${A} \in {\cal A}$ and ${B} \in {\cal B}$, the autoencoder transforms an input vector ${x} \in {\mathbb F}^n$ into an output vector ${A\circ B }({x}) \in {\mathbb F}^n$ (Figure \ref{fig:autoencoder}).
The corresponding {\it autoencoder problem} is to find ${ A} \in \cal A$ and ${ B} \in \cal B$ that minimize the overall distortion (or error/energy) function:

\be
\min E({A},{ B})=\min_{{ A},{B}} \sum_{t=1}^mE(x_t)=\min_{{ A},{B}} \sum_{t=1}^m \Delta \bigl ({A}\circ{ B}({ x}_t),{ x}_t \bigr)
\label{eq:objective1}
\ee
In the non auto-associative case, when external targets $y_t$ are provided, the minimization problem becomes:

\be
\min E({A},{ B})=\min_{{ A},{B}} \sum_{t=1}^mE(x_t)=\min_{{ A},{B}} \sum_{t=1}^m \Delta \bigl ({A}\circ{ B}({ x}_t),{ y}_t \bigr)
\label{eq:objective1a}
\ee

Note that $p<n$ corresponds to the regime where the autoencoder tries to implement some form of compression or feature extraction. The case
$p>n$ is not treated here but can be interesting in situations which either (1) prevent the use of trivial solutions by enforcing additional constraints, such as sparsity, or (2) include noise in the hidden layer, corresponding to transmission over a noisy channel.

Obviously, from this general framework, different kinds of autoencoders can be derived depending, for instance, on the choice of sets $\mathbb F$ and $\mathbb G$, transformation classes $\cal A$ and $\cal B$, distortion function $\Delta$,
as well as the presence of additional constraints. Linear autoencoders correspond to the case where $\mathbb F$ and $\mathbb G$ are fields and $\cal A$ and $\cal B$ are the classes of linear transformations, hence $A$ and $B$ are matrices of size $n \times p$ and $p \times n$ respectively. The linear real case where
$\mathbb F = \mathbb G =\mathbb R$ and $\Delta$ is the squared Euclidean distance was addressed in \cite{baldihornik88} (see also \cite{bourlard1988auto}).

\subsection{Complex Linear Autoencoder}
Here we consider the corresponding {\it complex} linear case where $\mathbb F = \mathbb G =\mathbb C$ and the goal is the minimization of the squared Euclidean distance

\be
\min E({A},{ B})=\min_{{ A},{B}} \sum_{t=1}^m \vert \vert x_t-AB(x_t) \vert\vert^2 = \sum_{t=1}^m  ( x_t-AB(x_t))^* (x_t-AB(x_t))
\label{eq:objective2}
\ee
Unless otherwise specified, all vectors are column vectors and we use $x^*$ (resp. $X^*$) to denote the conjugate transpose of a vector $x$ (resp. of a matrix $X$).
Note that the same notation works for both the complex and real case.
As we shall see, in the linear complex case as in the linear real case, one can also address the case where external targets are available, in which case the goal is the minimization of the distance

\be
\min E({A},{ B})=\min_{{ A},{B}} \sum_{t=1}^m \vert \vert y_t-AB(x_t) \vert\vert^2 = \sum_{t=1}^m  ( y_t-AB(x_t))^* (y_t-AB(x_t))
\label{eq:objective2-1}
\ee
In practical applications, it is often preferable to work with centered data, after substraction of the mean. The centered and non- centered versions of the problem are two different problems with in general two different solutions.
The general equations to be derived apply equally to both cases.

In general, we define the covariance matrices as follows

\be
\Sigma_{XY}=\sum_t x_ty^*_t
\label{eq:cov 1}
\ee
Using this definition, $\Sigma_{XX}, \Sigma_{YY}$ are Hermitian matrices $(\Sigma_{XX})^*=\Sigma_{XX}$ and $(\Sigma_{YY})^*=\Sigma_{YY}$, and $(\Sigma_{XY})^*=\Sigma_{ Y X}$.
We let also

\be
\Sigma=\Sigma_{YX}\Sigma_{XX}^{-1}\Sigma_{XY}
\label{eq:cov2}
\ee
$\Sigma$ is also Hermitian.
In the {\it auto-associative} case,  $x_t=y_t$ for all $t$ resulting in  $\Sigma=\Sigma_{XX}$.
Note that any Hermitian matrix admits a set of orthonormal eigenvectors and all its eigenvalues are real.
Finally, we let $I_m$ denote the $m \times m$ identity matrix.

For several results, we make the assumption that $\Sigma$ is invertible. This is not a very restrictive assumption for several reasons. First, by adding a small amount of noise to the data, a non-invertible $\Sigma$ could be converted to an invertible $\Sigma$, although this could potentially raise some numerical issues. More importantly, in most settings one can expect the training vectors to span the entire input space and thus $\Sigma$ to be invertible. If the training vectors span a smaller subspace, then the original problem can be transformed to an equivalent problem defined on the smaller subspace.

\subsection{Useful Reminders}

\noindent
{\bf Standard Linear Regression.} Consider the standard linear regression problem of minimizing
$E(B)=\sum_t \vert\vert y_t-Bx_t \vert\vert^2$, where $B$ is a $p\times n$ matrix, corresponding to a linear neural network without any hidden layers.
Then we can write

\begin{equation}
E(B)=\sum_t x_t^*B^*Bx_t-2{\rm Re}\, (y_t^*Bx_t)+||y_t||^2
\label{eq:fixed01}
\end{equation}
Thus $E$ is a convex function in $B$ because the associated quadratic form is equal to

\begin{equation}
\sum_t x_t^*C^*Cx_t=\sum_t||Cx_t||^2\geq 0
\label{eq:fixed02}
\end{equation}
Let $B$ be a critical point. Then by definition for any $p\times n$ matrix $C$ we must have
$lim_{\epsilon \to 0}$ $[E(B+\epsilon C) -E(B)]/\epsilon =0$. Expanding and simplifying this expression gives

\be
\sum_t x_t^*B^*Cx_t-y_t^*BC x_t=0
\label{eq:fixed03}
\ee
for all $p\times n$ matrices $C$. Using the linearity of the trace operator and its invariance under circular permutation of its arguments\footnote{It is easy to show directly that for any matrices $A$ and $B$ of the proper size, ${\rm Tr}(AB)={\rm Tr}(BA)$ \cite{lang1984algebra}. Therefore for any matrices $A$, $B$, and $C$ of the proper size, we have
${\rm Tr}(ABC)={\rm Tr}(CAB)={\rm Tr}(BCA)$.}, this is equivalent to

\be
{\rm Tr}\, ((\Sigma_{XX}B^*-\Sigma_{XY})C)=0
\label{eq:fixeda7-2}
\ee
for any $C$. Thus we have $\Sigma_{XX}B^*-\Sigma_{XY}=0$ and therefore

\be
B\Sigma_{XX}=\Sigma_{YX}
\label{eq:fixeda8}
\ee
If $\Sigma_{XX}$ is invertible, then $Cx_t=0$ for any $t$ is equivalent to $C=0$, and thus the function $E(B)$ is strictly convex in $B$. The unique critical point is the global minimum given by $B=\Sigma_{YX}\Sigma_{XX}^{-1}$.
As we shall see, the solution to the standard linear regression problem, together with the general approach given here to solve it, is also key for solving the more general linear autoencoder problem. The solution will also involve projection matrices.

\noindent
{\bf Projection Matrices.} For any $n\times k$ matrix $A$ with $k\leq n $, let $P_A$ denote the orthogonal projection onto the subspace generated by the columns of $A$. Then $P_A$ is  a Hermitian symmetric matrix and $P_A^2=P_A$, $P_AA=A$ since the image of $P_A$ is spanned by the columns of $A$ and these are invariant under $P_A$. The kernel of $P_A$ is the space $A^\perp$ orthogonal to the space spanned by the columns of $A$. Obviously, we have $P_AA^\perp=0$ and
$A^*P_A=A^*$. The projection onto the space orthogonal to the space spanned by the columns of $A$ is given by $I_n-P_A$.
In addition, if the columns of $A$ are independent (i.e. $A$ has full rank $k$),
then the matrix of the orthogonal projection is given by $P_A=A(A^*A)^{-1}A^*$
\cite{meyer2000matrix} and $P_A^*=P_A$.  Note that all these relationships are true even when the columns of $A$ are not orthonormal.

\subsection{Some Misconceptions}

As we shall see, in the complex case as in the real case, the global minimum corresponds to Principal Component Analysis. While the global minimum solution of linear autoencoders over infinite fields can be expressed analytically, it is often not well appreciated that there is more to be understood about linear autoencoders and the landscape of $E$. In particular, if one is interested in learning algorithms that proceed through incremental and somewhat ``blind'' weight adjustments, then one must study the entire landscape of $E$, including all the critical points of $E$, and derive and compare different learning algorithms. A second misconception is to believe that the problem is a convex optimization problem, hence somewhat trivial, since after all the error function is quadratic and the transformation $W=AB$ is linear. The problem with this argument is that the small layer of size $p$ forces $W$ to be of rank $p$ or less, and the set of matrices or rank at most $p$ is {\it not} convex. Furthermore, the problem is not convex when finite fields are considered. What is true and crucial for solving the linear autoencoders over infinite fields is that the problem becomes convex when $A$ or $B$ is fixed. A third misconception, related to the illusion of convexity, is that the $L_2$ landscape of linear neural networks never has any local minima. In general this is not true, especially if there are additional constraints on the linear transformation, such as restricted connectivity between layers so that some of the matrix entries are constrained to assume fixed values.

\section {Group Invariances}
For any autoencoder, it is important to investigate whether there are any group of transformations that leave its properties invariant.

\noindent
{\bf Change of Coordinates in the Hidden Layer.} Note that for any invertible $p \times p$ complex matrix $C$, we have $W=AB=ACC^{-1}B$ and $E(A,B)=E(AC,C^{-1}B)$. Thus all the properties of the linear autoencoder are fundamentally invariant with respect to any change of coordinates in the hidden layer.

\noindent
{\bf Change of Coordinates in the Input/Output Spaces.}
Consider an orthonormal change of coordinates in the output space defined by an orthogonal (or unitary) $n \times n$ matrix $D$, and any change of coordinates in the input space defined by an invertible $n\times n$ matrix $C$. This leads to a new autoencoder problem with input vectors  $Cx_1, \ldots, Cx_m$ and target output vectors of the form $Dy_1,\ldots, Dy_m$ with reconstruction error of the form

\be
E(A',B')=\sum_t \vert\vert Dy_t-A'B'Cx_t \vert\vert^2
\label{eq:group1}
\ee
If we use the one-to-one mapping between pairs of matrices $(A,B)$ and $(A',B')$ defined by $A'=DA$ and $B'=BC^{-1}$, we have

\be
E(A',B')=\sum_t \vert\vert Dy_t-A'B'Cx_t \vert\vert^2=\sum_t \vert\vert Dy_t-DABx_t \vert\vert^2=\sum_t \vert\vert y_t-ABx_t \vert\vert^2
\label{eq:group2}
\ee
the last equality using the fact that $D$ is an isometry which preserves distances.
Thus, using the transformation $A'=DA$ and $B'=BC^{-1}$
the original problem and the transformed problem are equivalent and the function $E(A,B)$ and $E(A',B')$ have the same landscape. In particular, in the auto-associative case, we can take $C=D$ to be a unitary matrix. This leads to an equivalent autoencoder problems with input vectors $Cx_t$ and covariance matrix $C\Sigma C^{-1}$. For the proper choice of $C$ there is an equivalent problem where basis of the space is provided by the eigenvectors of the covariance matrix and the covariance matrix is a diagonal matrix with diagonal entries equal to the eigenvalues of the original covariance matrix $\Sigma$.

\section{Fixed-Layer and Convexity Results}

A key technique for studying any autoencoder, is to simplify the problem by fixing all its transformations but one. Thus in this section we study what happens to the complex-valued linear autoencoder problem when either $A$ or $B$ is fixed, essentially reducing the problem to standard linear regression. The same approach can be applied to an autoencoder with more than one hidden layer (see section on Deep Architectures).

\begin{theorem} \label{thm1}
{\bf (Fixed A)}
For any fixed $n\times p$ matrix $A$,  the function $E(A,B)$ is convex in the coefficients of $B$ and attains its minimum for any $B$ satisfying the equation

\begin{equation}
A^*AB\Sigma_{XX}=A^*\Sigma_{YX}
\label{eq:fixeda1}
\end{equation}
If $\Sigma_{XX}$ is invertible and $A$ is of full rank $p$, then $E$ is strictly convex and has a unique minimum reached when

\begin{equation}
B=(A^*A)^{-1}A^*\Sigma_{YX}\Sigma_{XX}^{-1}
\label{eq:fixeda2}
\end{equation}
In the auto-associative case, if $\Sigma_{XX}$ is invertible and $A$ is of full rank $p$, then the optimal $B$ has full rank $p$ and does not depend on the data. It is given by
\begin{equation}
B=(A^*A)^{-1}A^*
\label{eq:fixeda3}
\end{equation}
and in this case, $W=AB=A(A^*A)^{-1}A^*=P_A$ and $BA=I_p$.
\end{theorem}

\noindent
{\bf Proof.} We write

\begin{equation}
E(A,B)=\sum_t x_t^*B^*A^*ABx_t-2{\rm Re}\, (y_t^*ABx_t)+||y_t||^2
\label{eq:fixeda4}
\end{equation}
Then for fixed $A$, $E$ is a convex function because the associated quadratic form is equal to

\begin{equation}
\sum_t x_t^*C^*A^*ACx_t=\sum_t||ACx_t||^2\geq 0
\label{eq:fixeda5}
\end{equation}
for any $p\times n$ matrix $C$.
Let $B$ be a critical point. Then by definition for any $p\times n$ matrix $C$ we must have
$lim_{\epsilon \to 0}$ $[E(A, B+\epsilon C) -E(A,B)]/\epsilon =0$. Expanding and simplifying this expression gives

\be
\sum_t x_t^*B^*A^*ACx_t-y_t^*AC x_t=0
\label{eq:fixeda6}
\ee
for all $p\times n$ matrices $C$. Using the linearity of the trace operator and its invariance under circular permutation of its arguments,
this is equivalent to

\be
{\rm Tr}\, ((\Sigma_{XX}B^*A^*A-\Sigma_{XY}A)C)=0
\label{eq:fixeda7}
\ee
for any $C$. Thus we have $\Sigma_{XX}B^*A^*A-\Sigma_{XY}A=0$ and therefore

\be
A^*AB\Sigma_{XX}=A^*\Sigma_{YX}
\label{eq:fixeda8-2}
\ee
Finally, if $\Sigma_{XX}$ is invertible and if $A$ is of full rank, then $ACx_t=0$ for any $t$ is equivalent to $C=0$, and thus the function $E(A,B)$ is strictly convex in $B$. Since $A^*A$ is invertible,  the unique critical point is obtained by solving Equation~\ref{eq:fixeda1}.

In similar fashion, we have the following theorem.

\begin{theorem} [{\bf Fixed B}]\label{thm2}

For any fixed $p\times n$ matrix $B$, the function $E(A,B)$ is convex in the coefficients of $A$ and attains its minimum for any $A$ satisfying the equation

\begin{equation}
AB\Sigma_{XX}B^*=\Sigma_{YX}B^*
\label{eq:fixedb1}
\end{equation}
If $\Sigma_{XX}$ is invertible and $B$ is of full rank, then $E$ is strictly convex and has a unique minimum reached when

\be
A=\Sigma_{YX}B^*(B\Sigma_{XX}B^*)^{-1}
\label{eq:fixedb2}
\ee
In the auto-associative case, if $\Sigma_{XX}$ is invertible and $B$ is of full rank, then the optimal $A$ has full rank $p$ and depends on the data. It is given by

\be
A=\Sigma_{XX}B^*(B\Sigma_{XX}B^*)^{-1}
\label{eq:fixedb3}
\ee
and $BA=I_p$.
\end{theorem}

\noindent
{\bf Proof.}
From Equation~\ref{eq:fixeda4}, the function $E(A,B)$ is a convex function in $A$.
The condition for $A$ to be a critical point is

\be
\sum_t x_t^*B^*A^*CBx_t-y_t^*CBx_t=0
\label{eq:fixedb4}
\ee
for any $p\times n$ matrix $C$, which is equivalent to

\be
{\rm Tr}\,((B\Sigma_{XX}B^*A^*-B\Sigma_{XY})C)=0
\label{eq:fixedb5}
\ee
for any matrix $C$. Thus $B\Sigma_{XX}B^*A^*-B\Sigma_{XY}=0$ which implies Equation~\ref{eq:fixedb1}. The other assertions of the theorem can easily be deduced.
\begin{remark}
Note that from Theorems 1 and 2 and their proofs, we have that $(A,B)$ is a critical point of $E(A,B)$ {\it if and only if} Equation \ref{eq:fixeda1}
and Equation \ref{eq:fixedb1} are simultaneously satisfied, that is if and only if $A^*AB\Sigma_{XX}=A^*\Sigma_{YX}$ and $AB\Sigma_{XX}B^*=\Sigma_{YX}B^*$.
\end{remark}
\section{Critical Points and the Landscape of $E$}

In this section we further study the landscape of $E$, its critical points, and the properties of $W=AB$ at those critical points.
\begin{theorem} {\bf (Critical Points)}
Assume that $\Sigma_{XX}$ is invertible. Then two  matrices $(A,B)$  define a critical point of $E$, if and only if the global map $W=AB$ is of the form

\begin{equation}
W=P_A\Sigma_{YX}\Sigma_{XX}^{-1}
\label{eq:critic1}
\end{equation}
with $A$ satisfying

\begin{equation}
P_A\Sigma=P_A\Sigma P_A=\Sigma P_A
\label{eq:critic2}
\end{equation}
In the auto-associative  case, the above becomes

\be
W=AB=P_A
\label{eq:critic3}
\ee
and
\be
P_A\Sigma_{XX}=P_A\Sigma_{XX}P_A=\Sigma_{XX} P_A
\label{eq:critic4}
\ee
If $A$ is of full rank, then the pair $(A, B)$ defines a critical point of $E$ if and only if $A$ satisfies
Equation~\ref{eq:critic2} and $B$ satisfies Equation~\ref{eq:fixeda3}. Hence $B$ must also be of full rank.
\end{theorem}

\noindent
{\bf Proof.} If $(A,B)$ is a critical point of $E$, then from Equation~\ref{eq:fixeda1}, we must have

\be
A^*(AB-\Sigma_{YX}\Sigma_{XX}^{-1})=0
\label{eq:critic5}
\ee
Let

\be
S=AB-P_A\Sigma_{YX}\Sigma_{XX}^{-1}
\label{eq:critic6}
\ee
Then since $A^*P_A=A^*$, we have $A^*S=0$. Thus the space spanned by the columns of $S$ is a subset of the space orthogonal to the space spanned by the columns of $A$ (i.e. $S \in A^{\perp}$). On the other hand, since

\be
P_AS=S
\label{eq:critic7}
\ee
$S$ is also in the space spanned by the columns of $A$ (i.e. $S \in Span(A)$).
Taken together, these two facts imply that $S=0$, resulting in $W=AB=P_A\Sigma_{YX}\Sigma_{XX}^{-1}$,
which proves Equation~\ref{eq:critic1}. Note that for this result, we need only $B$ to be critical (i.e. optimized with respect to $A$).
Using the definition of $\Sigma$, we have

\be
P_A\Sigma P_A=P_A\Sigma_{YX}\Sigma_{XX}^{-1}\Sigma_{XX}\Sigma_{XX}^{-1}\Sigma_{XY}P_A
\label{eq:critic8}
\ee
Since $S=0$, we have $AB=P_A\Sigma_{YX}\Sigma_{XX}^{-1}$ and thus

\be
P_A\Sigma P_A=P_A\Sigma_{YX}\Sigma_{XX}^{-1}\Sigma_{XX}\Sigma_{XX}^{-1}\Sigma_{XY}P_A=AB\Sigma_{XX}B^*A^*
\label{eq:critic81}
\ee
Similarly, we have

\be
P_A\Sigma=AB\Sigma_{XY}
\label{eq:critic9}
\ee
and

\be
\Sigma P_A=\Sigma_{YX}B^*A^*
\label{eq:critic10}
\ee
Then Equation~\ref{eq:critic2} result immediately by combining Equations \ref{eq:critic81}, \ref{eq:critic9}, and \ref{eq:critic10} using Equation~\ref{eq:fixedb1}.
The rest of the theorem follows easily.
\begin{remark} The above proof unifies the cases when $AB$ is of rank $p$ and less than $p$ and avoids the need for two separate proofs, as was done in earlier work \cite{baldihornik88} for the real-valued case.
\end{remark}
\noindent

\begin{theorem} {\bf (Critical Points of Full Rank)} \label{thm4}
Assume that $\Sigma$ is of full rank with $n$ distinct eigenvalues $\lambda_1>\cdots>\lambda_n$ and let $u_1,\ldots,u_n$ denote a corresponding basis of orthonormal eigenvectors. If ${\mathcal I}=\{i_1,\ldots,i_p\}$ $(1 \leq i_1< \ldots < i_p \leq n)$ is any ordered set of indices of size $p$, let
 $U_{\mathcal I}=(u_{i_1}, \ldots,u_{i_p})$ denote the matrix formed using the corresponding column eigenvectors.
 Then two full rank matrices $A,B$ define a critical point of $E$ if and only if there exists an ordered $p$-index set $\mathcal I$ and an invertible $p\times p$ matrix $C$ such that

\be
A=U_{\mathcal I}C \quad {\rm and} \quad B=C^{-1}U^*_{\mathcal I}\Sigma_{YX}\Sigma_{XX}^{-1}
\label{eq:global1}
\ee
For such critical point, we have

\be
W=AB=P_{U_{\mathcal I}} \Sigma_{YX} \Sigma_{XX}^{-1}
\label{eq:global2}
\ee
and

\be
E(A,B)={\rm Tr} \,\Sigma_{YY}- \sum_{i \in {\mathcal I}}\lambda_i
\label{eq:global3}
\ee
In the auto-associative case, these equations reduce to

\be
A=U_{\mathcal I}C \quad {\rm and} \quad B=C^{-1}U^*_{\mathcal I}
\label{eq:global4}
\ee

\be
W=AB=P_{U_{\mathcal I}}
\label{eq:global5}
\ee
and

\be
E(A,B)={\rm Tr}\, \Sigma- \sum_{i \in {\mathcal I}}\lambda_i=\sum_{i \in \bar {\mathcal I}}\lambda_i
\label{eq:global6}
\ee
where $\bar {\mathcal I}=\{1,\ldots, n\}\backslash{\mathcal I}$ is the complement of $\mathcal I$.
\end{theorem}

\par
\noindent
{\bf {Proof.}}
Since $P_A\Sigma=\Sigma P_A$, we have

\be
P_A\Sigma A=\Sigma P_AA=\Sigma A
\label{eq:global7}
\ee
Thus the columns of $A$ form an invariant space of $\Sigma$. Thus $A$ is of the form $U_{\mathcal I}C$.
The conclusion for $B$  follows from Equation~\ref{eq:critic1} and the rest is easily deduced, as in the real case. Equation \ref{eq:global6} can be derived easily by using the remarks in Section 3 and using the unitary change of coordinates under which $\Sigma_{XX}$ becomes a diagonal matrix. In this system of coordinates,
we have
\[
E(A,B)=\sum_t ||y_t||^2+\sum_t {\rm Tr}\, (x_t^*(AB)^*AB x_t)-2\sum_t {\rm Tr}\, (y_t^*ABx_t)
\]
Therefore, using the invariance property of the trace under permutation, we have
\[
E(A,B)={\rm Tr}\,(\Sigma)+{\rm Tr}\, ((AB)^2\Sigma)-2{\rm Tr}\, (AB\Sigma)
\]
Since $AB$ is a projection operator, this yields  Equation~\ref{eq:global6}.
In the auto-associative case with these coordinates it is easy to see that
$W(x_t)$ and $E(A,B)=\sum_t E(x_t)$  are easily computed from the values of $W(u_i)$. In particular, $E(A,B)=\sum_{i=1}^n \lambda_i (u_i-W(u_i))^2$.
In addition, at the critical points, we have $W(u_i)=u_i$ if $i\in I$, and $W(u_i)=0$ otherwise.

\begin{remark}
All the previous theorems are true in the hetero-associative case with targets $y_t$. Thus they can readily be applied to address the linear denoising autoencoder \cite{vincent2008extracting,vincent2011connection} over $\mathbb R$ or $\mathbb C$. The linear denoising autoencoder is an autoencoder trained to remove noise by having to associate noisy versions of the inputs with the correct inputs. In other words, using the current notation, it is an autoencoder where the inputs $x_t$ are replaced by $x_t+n_t$ where $n_t$ is the noise vector and the target outputs $y_t$ are of the form $y_t=x_t$. Thus the previous theorems can be applied using the following replacements: $\Sigma_{XX}=\Sigma_{XX}+\Sigma_{NN}+\Sigma_{NX}+\Sigma_{XN}$, $\Sigma_{XY}=\Sigma_{XX}+\Sigma_{NX}$, $\Sigma_{YX}=\Sigma_{XX}+\Sigma_{XN}$. Further simplifications can be obtained using particular assumptions on the noise, such as
$\Sigma_{NX}=\Sigma_{XN}=0$.
\end{remark}

\begin{theorem} {\bf (Absence of Local Minima)}
The global minimum of the complex linear autoencoder is achieved by full rank matrices $A$ and $B$ associated with the index set $1, \ldots, p$ of the $p$ largest eigenvalues of $\Sigma$ with $A=U_{\mathcal I}C$ and $B=C^{-1}U^*_{\mathcal I}$ (and where $C$ is any invertible $p \times p$ matrix). When $C=I$, $A=B^*$. All other critical points are saddle points associated with corresponding projections onto non-optimal sets of eigenvectors of $\Sigma$ of size $p$ or less.
\end{theorem}

\noindent
{\bf {Proof.}} The proof is by a perturbation argument, as in the real case, showing that critical points that are not associated with the global minimum there is always a direction of escape that can be derived using unused eigenvectors associated with higher eigenvalues in order to lower the error $E$ (see \cite{baldihornik88} for more details). The proof can be made very simple by using the group invariance properties under transformation of the coordinates by a unitary matrix. With such a transformation, it is sufficient to study the landscape of $E$ when $\Sigma$ is a diagonal matrix and $A=B^*=U_{\cal I}$.

\begin{remark}
At the global minimum, if $C$ is the $p \times p$ identity matrix ($C=I$), in the auto-associative case then the activities in the hidden layer are given by
$u_1^*x, \ldots, u_p^*x$, corresponding to the coordinates of $x$ along the first $p$ eigenvectors of $\Sigma_{XX}$. These are the so called principal components of $x$ and the autoencoder implements a form of Principal Component Analysis (PCA) also closely related to Singular Value Decomposition (SVD).
\end{remark}

\begin{figure}[ht]
\begin{center}
\includegraphics[width=0.80\columnwidth]{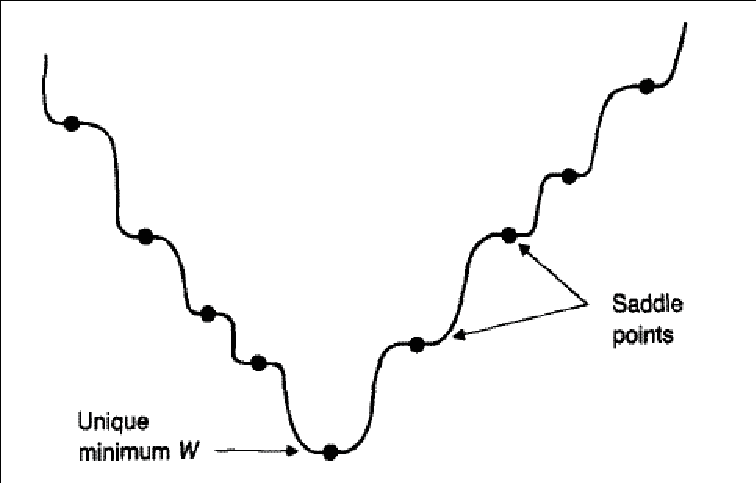}
\end{center}
\caption{Landscape of $E$.}
\label{fig:autoencoder-2}
\end{figure}

The theorem above shows that when $\Sigma$ is full rank, there is a special class of critical points associated with $C=I$. In the auto-associative case, this class is characterized by the fact that $A$ and $B$ are conjugate transpose of each other ($A=B^*$) in the complex-valued case, or transpose of each other ($A=B^*$) in the real-valued case. This class of critical points is special for several reasons. For instance, in the related Restricted Boltzmann Machine Autoencoders the weights between visible and hidden units are require to be symmetric corresponding to $A=B^*$.
More importantly, these critical points are closely connected to Hebbian learning (see also \cite{oja1982simplified,oja1989neural,oja1992principal}). In particular,
for linear real-valued autoencoders, if $A=B^*$ and $E=0$ so that inputs are equal to outputs, any learning rule that is symmetric with respect to the pre- and post- synaptic activities--which is typically the case for Hebbian rules--will modify $A$ and $B$ but preserve the property that $A=B^*$. This remains roughly true even if $E$ is not exactly zero. Thus for linear real-valued autoencoders, there is something special about transposition operating on $A$ and $B$ and more generally on can suspect a similar role is played by conjugate transposition in the case of linear complex-valued autoencoders. The next theorem and the following section on learning algorithm further clarify this point.

\begin{theorem} {\bf (Conjugate Transposition)}
Assume $\Sigma_{XX}$ is of full rank in the auto-associative case.
Consider any point $(A,B)$ where $B$ has been optimized with respect to $A$, including all critical points. Then

\be
W=AB=B^*A^*AB=B^*A^*=W^* \quad {\rm and} \quad E(A,B)=E(B^*,A^*)
\label{eq:hebb100}
\ee
Furthermore, when $A$ is full rank

\be
W=P_A=P_A^*=W^*
\label{eq:hebb1000}
\ee
\end{theorem}

\noindent
{\bf Proof.}
By Theorem 1, in the auto-associate case, we have
\[
A^*AB=A^*
\]
Thus, by taking the complex conjugate of each side, we have
\[
B^*A^*A=A
\]
It follows that
\[
B^*A^*=B^*A^*AB=AB
\]
which proves Equation \ref{eq:hebb100}. If in addition $A$ is full rank, then by Theorem 1 $W=AB=P_A$ and the rest follows immediately.

\begin{remark}
Note the following. Starting from a pair $(A,B)$ with $W=AB$ and where $B$ has been optimized with respect to $A$, let $A'=B^*$ and optimize $B$ again so that $B'=(A'{A'}^*)^{-1}{A'}^*$. Then we also have

\be
W'=A'B'=W^* =W=P_A \quad {\rm and} \quad E(A,B)=E(A',B')
\label{eq:inter2}
\ee
\end{remark}

\section{Optimization or Learning Algorithms}

Although mathematical formula for the global minimum solution of the linear autoencoder have been derived,
the global solution may not be available immediately to a self-adjusting learning circuit capable of making only small adjustments at each learning steps. Small adjustments may also be preferable in a non-stationary environment where the set $\cal X$ of training vectors changes with time.
Furthermore, the study of small adjustment algorithms in linear circuits may shed some light on
similar incremental algorithms applied to non-linear circuits where the global optimum cannot be derived analytically.
Thus, from a learning algorithm standpoint, it is still useful to consider incremental optimization algorithms, such as gradient descent or partial EM steps, even when such algorithms are slower or less accurate than direct global optimization.
The previous theorems suggest two kinds of operations that could be used in various combinations to iteratively minimize $E$, taking full or partial steps: (1) Partial minimization: fix $A$ (resp. $B$) and minimize for $B$ (resp. $A$); (2) Conjugate Transposition: fix $A$ (resp. $B$), and set $B=A^*$ (resp. $A=B^*)$ (the latter being reserved for the auto-associative case, and particularly so if one is interested in converging to solutions where $A$ and $B$ are conjugate transpose of each other, i.e. where $C=I$).

\begin{theorem} {\bf (Alternate Minimization)}
Consider the algorithm where $A$ and $B$ are optimized in alternation (starting from $A$ or $B$), holding the other one fixed. This algorithm will converge to a critical point of $E$. Furthermore, if the starting value of $A$ or $B$ is initialized randomly, then with probability one the algorithm will converge to a critical point where
both $A$ and $B$ are full rank.
\end{theorem}

\noindent
{\bf {Proof:}} A direct proof of convergence is given in Appendix B. Here we give an indirect, but perhaps more illuminating proof, by remarking that the alternate minimization algorithm is in fact an instance of the general EM algorithm \cite{dempster77} combined with a hard decision, similar to the Viterbi learning algorithm for HMM or the k-means clustering algorithm with hard assignment. For this,
consider that we have a probabilistic model over the data with parameters $A$ and hidden variables $B$, or vice versa, with parameters $B$ and hidden variables $A$. The conditional probability of the data and the hidden variables is given by:

\be
P({\cal {X}},{\cal {Y}}, A|B)={1 \over Z_1} e^{-E(A,B)}
\label{eq:probamodel}
\ee
or

\be
P({\cal {X}},{\cal {Y}}, B|A)={1 \over Z_2} e^{-E(A,B)}
\label{eq:probamodel-2}
\ee
where $Z1$ and $Z_2$ denote the proper normalizing constants (partition functions).
During the E step, we find the most probable value of the hidden variables given the data and current value of the parameters. Since $E$ is quadratic, the model in Equation \ref{eq:probamodel}
is Gaussian and the mean and mode are identical. Thus the hard assignment of the hidden variables in the E step corresponds to optimizing $A$ or $B$ using Theorem 3 or Theorem 4. During the M step, the parameters are optimized given the value of the hidden variables. Thus the M step also
corresponds to optimizing $A$ or $B$ using Theorem 3 or Theorem 4. As a result, convergence to a critical point of $E$ is ensured by the general convergence theorem of the EM algorithm \cite{dempster77}.
Since $A$ and $B$ are initialized randomly, they are full rank with probability one and, by Theorem 1 and 2 they retain their full rank after each optimization step. Note that the error $E$ is always positive, strictly convex in $A$ or $B$, decreases at each optimization step, and thus $E$ must converge to a limit.
By looking at every other step in the algorithm, it is easy to see that $P_A$ must converge. From which one can see that $A$ must converge, and so must $B$.

\begin{figure}[!h]
\begin{center}
\includegraphics[width=6.5cm]{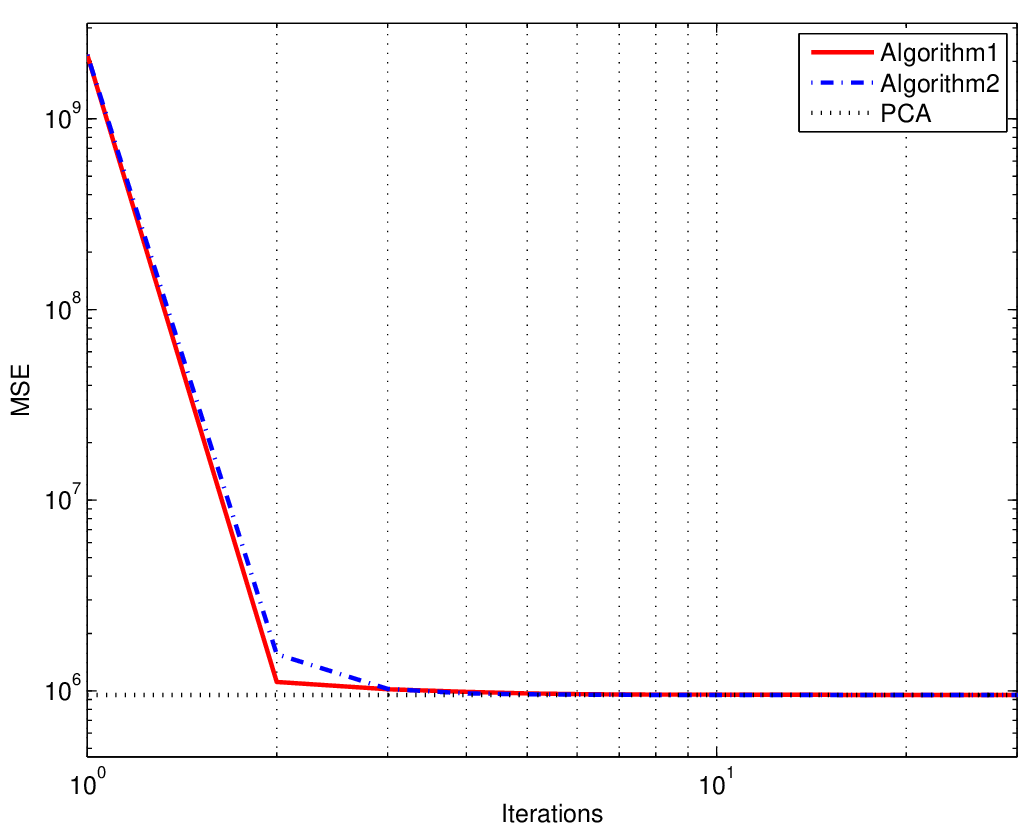}
\vspace{0.0in}
\includegraphics[width=6.5cm]{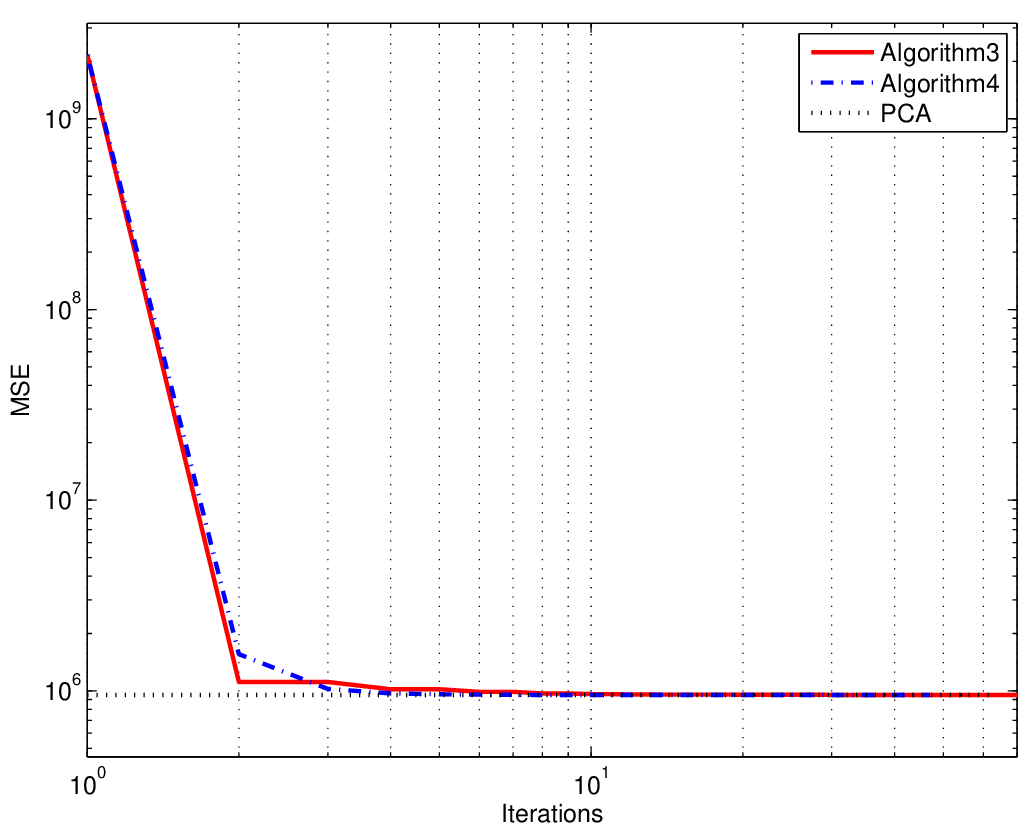}
\vspace{0.0in}
\includegraphics[width=6.5cm]{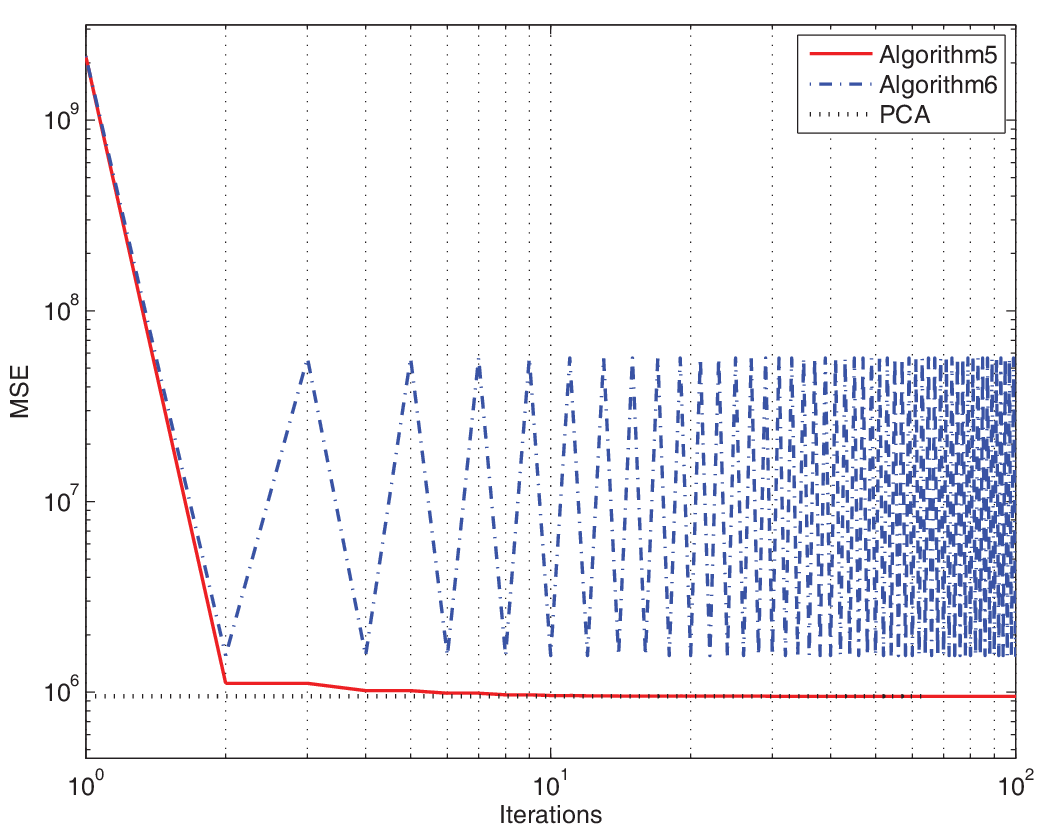}
\end{center}
\caption{Learning Curves for Algorithms 1-6.
The results are obtained using linear real-valued autoencoders of size 784-10-784  trained on images in the standard MNIST dataset for the digit ``7'' using 1,000 samples. Each consecutive update of both $A$ and $B$ is considered as one iteration.
\label{fig:plots}}
\end{figure}

Given the importance of conjugate transposition (Theorem 6) in the auto-associative case, one may also consider algorithms where the operations of conjugate transposition and partial optimization of $A$ and $B$ are interleaved.
This can be carried in many ways.
Let $A \longrightarrow B$ denote that $B$ is obtained from $A$ by optimization (Equation \ref{eq:fixeda3})
and $A \Longrightarrow B$ denote that $B$ is obtained from $A$ by conjugate transposition ($B=A^*$), and similarly for
 $B \longrightarrow A$ (Equation \ref{eq:fixedb3}) and  $B \Longrightarrow A$ ($A=B^*$).
 Let also $\Longleftrightarrow$ denote the operation where both $A$ and $B$ are obtained by simultaneous conjugate transposition from their current values. Then starting from (random) $A$ and $B$, here are several possible algorithms:

\begin{itemize}
\item {\bf Algorithm 1:} $B \longrightarrow A \longrightarrow B \longrightarrow A \longrightarrow B \ldots$.
\item {\bf Algorithm 2:} $A \longrightarrow B \longrightarrow A \longrightarrow B \longrightarrow A \ldots$.
\item {\bf Algorithm 3:} $B \longrightarrow A \longrightarrow B \Longrightarrow A \longrightarrow B \longrightarrow A \longrightarrow B \Longrightarrow A \ldots$.
\item {\bf Algorithm 4:} $A \longrightarrow B \longrightarrow A \Longrightarrow B \longrightarrow A \longrightarrow B \longrightarrow A \Longrightarrow B \ldots $.
\item {\bf Algorithm 5:} $B \longrightarrow A \longrightarrow B  \Longleftrightarrow  B \longrightarrow A \longrightarrow B \ldots$.
\item {\bf Algorithm 6:} $A \longrightarrow B \longrightarrow A \Longleftrightarrow  A \longrightarrow B \longrightarrow A \Longleftrightarrow \ldots$.
    \item {\bf Algorithm 7:} $A \longleftarrow B \Longleftrightarrow A \longleftarrow B \Longleftrightarrow \ldots$.
\end{itemize}
The theory presented so far allows us to understand their behavior easily (Figure \ref{fig:plots}), considering
a consecutive update of $A$ and $B$ as one iteration.
Algorithms 1 and 2 converge with probability one to a critical point where $A$ and $B$ are full rank.
Algorithm 1 may be slightly faster than Algorithm 2 at the beginning since in the first step Algorithm 1 takes into account the data (Equation \ref{eq:fixedb3}), whereas Algorithm 2 ignores it.
Algorithms 3, 4, and 5 converge and lead to a solution where $A=B^*$ (or, equivalently, $C=I$). Algorithms 3 and 5 take the same time and are faster than Algorithm 4. Algorithm 2 and Algorithm 4 take the same time.
Algorithm 3 requires almost twice the number of steps of Algorithm 1.
But Algorithm 4 is faster than Algorithm 3. This is because in Algorithm 3, the steps $B \Longrightarrow A \longrightarrow B$
is basically like switching the matrices $A$ and $B$, and the error after the step $B \longrightarrow A \longrightarrow B$ is the same as the error after the step $B\Longrightarrow A \longrightarrow B$.
Algorithms 6 and 7 in general will not converge. Only optimization steps with respect to the $B$ matrix are being carried and therefore the data is never considered.

\section{Generalization Properties}

One of the most fundamental problems in machine learning is to understand the generalization properties of a learning system. Although in general this is not a simple problem, in the case of the autoencoder the generalization properties can easily be understood. After learning, $A$ and $B$ must be at a critical point. Assuming without much loss of generality that $A$ is also full rank and $\Sigma_{XX}$ is invertible, then from Theorem 1 we know in the auto-associative case that $W=P_A$. Thus we have the following result.

\begin{theorem} {\bf (Generalization Properties)}
Assume in the auto-associative case that $\Sigma_{XX}$ is invertible. For any learning algorithm that converges to a point where $B$ is optimized with respect to $A$ and $A$ is full rank (including all full rank critical points), then for any vector $x$ we have $Wx=ABx=P_{A}x$ and

\be
E(x)=\vert\vert x-ABx\vert\vert^2=\vert\vert x -P_Ax \vert\vert^2
\label{eq:gene1}
\ee
\end{theorem}

\begin{remark}
Thus the reconstruction error of any vector is equal to the square of its distance to the subspace spanned by the columns of $A$, or the square of the norm of its projection onto the orthogonal subspace.
The general hetero-associative case can also be treated using Theorem 1. In this case, under the same assumptions,
we have: $W=P_A \Sigma_{YX} \Sigma_{XX}^{-1}$.
\end{remark}

\section{Recycling or Iteration Properties}

Likewise, for the linear auto-associative case, one can also easily understand what happens when the outputs of the network are recycled into the inputs after learning. In the RBMs case, this is similar to alternatively sampling from the input and hidden layer. Interestingly, this provides also an alternative characterization of the critical points.  At a critical points where $W$ is a projection, we must have $W^2=W$. Thus, after learning,
the iterates $W^mx$ are easy to understand and converge after a single cycle and
all points become stable after a single cycle.
If $x$ is in the space spanned by the columns of $A$ we have $W^m(x)=x$ for any $m \geq 1$. If $x$ is not in the space spanned by the columns of $A$, then $W^mx=y$ for $m \geq 2$, where $y$ is the projection of $x$ onto the space spanned by the columns of $A$ ($Wx=P_Ax=y$).

\begin{theorem} {\bf (Generalization Properties)}
Assume in the auto-associative case that $\Sigma_{XX}$ is invertible. For any learning algorithm that converges to a point where $B$ is optimized with respect to $A$ and $A$ is full rank (including all full rank critical points), then for any vector $x$ and any integer $m>1$, we have

\be
W^m(x)=P_A^m(x)=P_A(x)
\label{eq:gene1-2}
\ee
\end{theorem}

\begin{remark}
There is a partial converse to this result, in the following sense. Assume that $W$ is a projection ($W^2=W$) and therefore $ABAB=AB$. If $A$ is of full rank, then $BAB=B$. Furthermore, if $B$ is of full rank, then $BA=I_p$ (note that $BA=I_p$ immediately implies that $W^2=W$). Multiplying this relation by $A^*A$ on the left and $A$ on the right, yields $A^*AB=A^*$ after simplification, and therefore $B=(A^*A)^{-1}A^*$ Thus according to Theorem 1 $B$ is critical and $W=P_A$. Note that under the sole assumption that $W$ is a projection, there is no reason for $A$ to be critical, since there is no reason for $A$ to depend on the data and on $\Sigma_{XX}$.
\end{remark}

\section{Deep Architectures}

\begin{figure}[ht]
\begin{center}
\includegraphics[width=0.80\columnwidth]{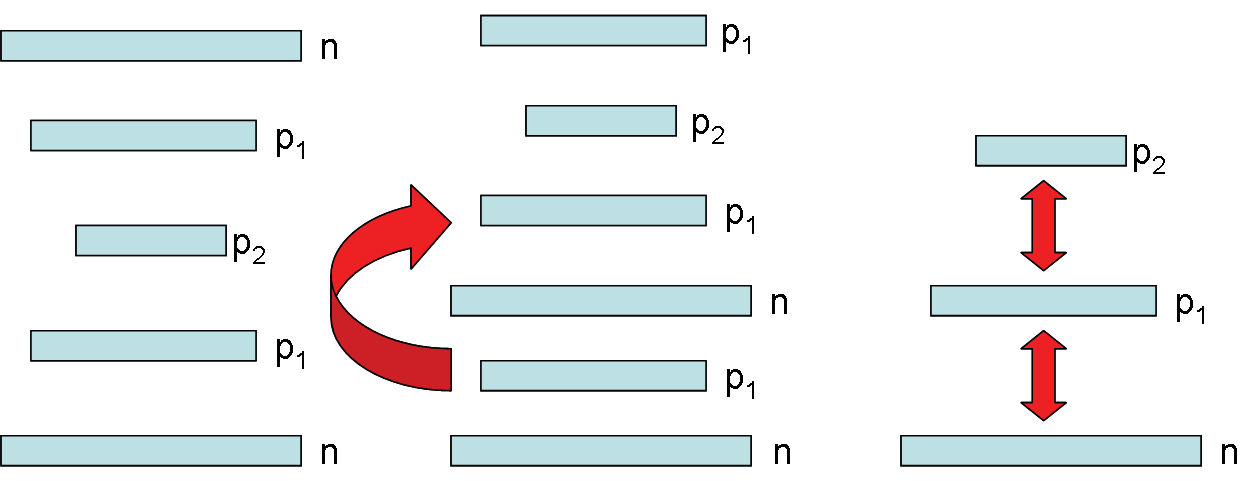}
\end{center}
\caption{Vertical Composition of Autoencoders.}
\label{fig:vertical}
\end{figure}

Autoencoders can be composed vertically (Figure \ref{fig:vertical}), as in the deep architecture approach described in
\cite{hinton2006fast,hinton2006reducing}, where a stack of RBMs is trained in an unsupervised way, in bottom up fashion,
by using the activity in the hidden layer of a RBM in the stack as the input for the next RBM in the stack.
Similar architectures and algorithms can be applied to linear networks.
 Consider for instance training a 10/5/10 autoencoder and then using the activities in the hidden layer to train a 5/3/5 autoencoder. This architecture can be contrasted with a 10/5/3/5/10 architecture, or a 10/3/10 architecture.
 In all cases, the overall transformation $W$ is linear and constrained in rank by the size of the smallest layer in the architecture. Thus all three architectures have the same optimal solution associated with Principal Component Analysis using the top 3 eigenvalues. However the landscapes of the error functions and the learning trajectories may be different and other considerations may play a role in the choice of an architecture.

In any case, the theory developed here can be adapted to multi-layer real-valued or complex-valued linear networks. Overall, such networks implement a linear transformation with a rank restriction associated with the smallest hidden layer. As in the single hidden layer case, the overall distortion is convex in any single matrix while all the other matrices are held fixed. Any algorithm that successively, or randomly, optimizes each matrix with respect to all the others will converge to a critical point, which will be full rank with probability one if the matrices are initialized randomly.
For instance, to be more precise, consider a network with five stages associated with the five matrices $A, B, C, D$ and $F$ of the proper sizes and the error function $E(A, B, C, D, F)=\sum_t \vert\vert (y_t-ABCDFx_t)\vert\vert^2$.

\begin{theorem}
For any fix set of matrices $A, B, D$ and $F$, the function $E(A, B, C, D, F)$ is convex in the coefficients of $C$ and attains its minimum for any $C$ satisfying the equation

\be
B^*A^*AB C DF\Sigma_{XX} F^*D^*=B^*A^*\Sigma_{YX}F^*D^*
\label{eq:deep1}
\ee
If $\Sigma_{XX}$ is invertible and $AB$ and $DF$ are of full rank, then $E$ is strictly convex and has a unique minimum reached when

\be
C=(B^*A^*AB)^{-1}B^*A^*\Sigma_{YX}F^*D^*(DF\Sigma_{XX} F^*D^*)^{-1}
\label{eq:deep2}
\ee
\end{theorem}

\noindent
{\bf Proof:} We write
\begin{equation}
E(A,B)=\sum_t x_t^*F^*D^*C^*B^*A^*ABCDFx_t-2{\rm Re}\, (y_t^*ABCDFx_t)+||y_t||^2
\label{eq:deep3}
\end{equation}
Then for fixed $A,B, D, F$, $E$ is a convex function because the associated quadratic form is equal to

\begin{equation}
\sum_t x_t^*F^*D^*L^*B^*A^*ABLDFx_t=\sum_t||ABLDFx_t||^2\geq 0
\label{eq:deep4}
\end{equation}
for any matrix $L$ of the proper size.
Let $C$ be a critical point. Then by definition for any matrix $L$ of the proper size, we must have
$lim_{\epsilon \to 0}$ $[E(A,B, C+\epsilon L,D,F) -E(A,B,C,D,F)]/\epsilon =0$. Expanding and simplifying this expression gives

\be
\sum_t x_t^*F^*D^*C^*B^*A^*ABLDFx_t-y_t^*ABLDF x_t=0
\label{eq:deep5}
\ee
for all matrices $C$ of the proper size. Using the linearity of the trace operator and its invariance under circular permutation of its arguments,
this is equivalent to

\be
{\rm Tr}\, ((DF\Sigma_{XX}F^*D^*C^*B^*A^*AB-DF\Sigma_{XY}AB)L)=0
\label{eq:deep6}
\ee
for any $L$. Thus we have $DF\Sigma_{XX}F^*D^*C^*B^*A^*AB-DF\Sigma_{XY}AB=0$ and therefore

\be
B^*A^*AB C DF\Sigma_{XX} F^*D^*=B^*A^*\Sigma_{YX}F^*D^*
\label{eq:deep7}
\ee
Finally, if $\Sigma_{XX}$ is invertible and $AB$ and $DF$ are of full rank,
then $ABLDFCx_t=0$ for any $t$ is equivalent to $L=0$, and thus the function $E(A,B,C,D,F)$ is strictly convex in $C$. Thus in this case we can solve Equation \ref{eq:deep7} for $C$ to get Equation \ref{eq:deep2}.

\section{Conclusion}

We have provided a fairly complete and general analysis of complex-valued linear autoencoders.
The analysis can readily be applied to special cases, for instance when the vectors are real-valued and the matrices are complex-valued, or the vectors are complex-valued and the matrices are real-valued. More importantly,
the analysis provides a unified view of real-valued and complex-valued linear autoencoders. In the Appendix, we further extend the treatment of linear autoencoders over infinite fields by looking at their properties from a
differential geometry perspective.

More broadly, the framework used here identifies key questions and strategies that ought to be studied for any class of autoencoders, whether linear or non-linear. For instance:
\begin{enumerate}
\item What are the relevant group actions and invariances for the problem?
\item Can one of the transformations ($A$ or $B$) be solved while the other is held fixed? Are there useful convex relaxations or restrictions?
\item Are there any critical points, and how can they be characterized?
\item Is there a notion of symmetry or transposition between the transformations $A$ and $B$ around critical points?
\item Is there an overall analytical solution? Is the problem NP-hard? What is the landscape of $E$?
\item What are the learning algorithms and their properties?
\item What are the generalization properties?
\item What happens if the outputs are recycled?
\item What happens if autoencoders are stacked vertically?
\end{enumerate}
All these questions can be raised anew for other linear autoencoders, for instance over $\mathbb R$ or $\mathbb C$ with the $L_p$ norm ($p\not = 2$), or over other fields, in particular over finite fields with the Hamming distance.
While results for finite fields will be published elsewhere, it is clear that these questions have different answers in the finite field case. For instance, the notion of using convexity to analytically solve for $A$ or $B$, while holding the other one fixed, breaks down in the finite field case.

These questions can also be applied to non-linear autoencoders. While in general non-linear autoencoders are difficult to treat analytically, the case of Boolean autoencoders was recently solved using this framework
\cite{baldijmlr11}. Boolean autoencoders implement a form of clustering when $p<n$ and, in retrospect, all linear autoencoders implement also a form of clustering when $p<n$. In the linear case, for any vector $x$ and any $W=AB$, we have $W(x+Ker W)=W(x)$. $Ker W$ is the kernel of $W$ which contains the kernel of $B$, and is equal to it when $A$ is of
full-rank. Thus, in general, linear autoencoders implement clustering ``by hyperplane'' associated with the kernel of $B$. Taken together, these facts point to the more general unity connecting unsupervised learning, clustering, Hebbian learning, and autoencoders.

Finally, there is the case of autoencoders,linear or non-linear, with $p \geq n$ which has not been addressed here.
Clearly, additional restrictions or conditions must be imposed in this case, such as sparse encoding in the hidden layer or sparse matrices using L1 regularization, to avoid trivial solutions associated with the identity function. Although beyond the scope of this paper, these autoencoders are also of interest. For instance, the linear case over finite fields with noise added to the hidden layer, subsumes the theory of linear codes in coding theory \cite{mceliece77}. Thus, in short, one can expect autoencoders to continue to play an important role in machine learning and provide fertile connections to other areas, from clustering to information and coding theory.

\section*{Appendix A: Differential Geometry of Autoencoders}

Methods from differential geometry has been applied effectively to statistical machine learning in previous studies by Amari \cite{amari1990differential,amari2007methods} and others. Here however we introduce a novel approach for looking at the manifolds of relevant parameters for linear autoencoders over the real or complex fields.
While the basic results in this section are not difficult, they do assume some understanding of the most basic concepts of differential geometry
\cite{spivak1999comprehensive}.

Let $R_p$ be the set of  $n\times n$ complex matrices of rank at most equal to $p$. Obviously, $AB\in R_p$.
In general, $R_p$ is a singular variety (a Brill-Noether variety).
We let also $R_p\backslash R_{p-1}$ be the set of $n\times n$ matrices of rank exactly $p$.
As we shall see,  $R_p\backslash R_{p-1}$ is a complex manifold.

\begin{definition}
We let

\be
F_p(W)=\sum_{t=1}^m||y_t-Wx_t||^2
\label{eq:F1}
\ee
where $W\in R_p$.
\end{definition}

\noindent
Let $M^{p \times q }$ be the set of all $p \times q$ complex matrices. Define the mapping

\be
\iota:M^{n\times p}\times M^{p\times n}\to R_p \quad {\rm with} \quad \iota(A,B)=AB
\label{eq:F2}
\ee
by taking the product of the corresponding matrices.
Then we have $F\circ\iota=E$. We are going to show that $\iota$ is surjective and the differential of $\iota$ is of full rank at any point.

\begin{lemma}
$R_p\backslash R_{p-1}$ is a complex manifold of dimension $2np-p^2$.
\end{lemma}

\noindent
{\bf Proof.} Let $W\in R_p\backslash R_{p-1}$. To construct a set of local coordinates of $R_p\backslash R_{p-1}$ near $W$, we write $W$

\be
W=(w_1,\cdots, w_n)
\label{eq:dim1}
\ee
where $w_1,\cdots,w_n$ are column vectors. Without any loss of generality, we assume that $ w_1,\cdots, w_p$ are linearly independent. Thus we must have

\be
w_j=\sum_{i=1}^p \xi_{ij} w_i
\label{eq:dim2}
\ee
for $j>p$, with complex coefficients $\xi_{ij}$. The local coordinates of $R_p\backslash R_{p-1}$ are
$(\xi_{ij})_{1\leq i\leq p,p<j\leq n}$ and $(w_{ik})_{1\leq i\leq p,1\leq k\leq n}$. Thus

\be
\dim (R_p\backslash R_{p-1})=p(n-p)+pn=2pn-p^2
\label{eq:dim3}
\ee

Next, we consider the tangent space $T_W$ of $R_p\backslash R_{p-1}$ at $W$. By
definition, a basis of
$T_C(R_p\backslash R_{p-1})$ is given by

\begin{align}
\frac{\pa}{\pa w_{ik}}& \qquad 1\leq i\leq p, 1\leq k\leq n;\\
\frac{\pa}{\pa \xi_{ij}}&\qquad 1\leq i\leq p, p<j\leq n.
\label{eq:tangent1}
\end{align}

Let $(e_1,\cdots, e_n)$ be the standard basis of ${\mathbb C}^n$. Then the corresponding matrices of the tangent vectors are

\begin{align*}
\frac{\pa}{\pa c_{ik}}& \longrightarrow (0,\cdots,\underset{ i-th\, place}{e_k},\cdots,0,\xi_{i,p+1}e_k,\cdots, \xi_{in}e_k);\\
\frac{\pa}{\pa \xi_{ij}}& \longrightarrow (0,\cdots,0,0,\cdots,\underset{j-th\, place}{w_i},\cdots,0).
\label{eq:tangent2}
\end{align*}

\vspace{.2in}

\begin{lemma} Let $W=AB$, where $A,B$ are full-rank $n\times p$ and $p\times n$ matrices, respectively. Let $A_1,B_1$ be $n\times p$ and $p\times n$ matrices such that
\be
AB_1+A_1B=0
\label{eq:tangent3}
\ee
Then there is an invertible $p\times p$ matrix $V$ such that

\be
A_1=AV, B_1=-VB
\label{eq:tangent4}
\ee
\end{lemma}

\noindent
{\bf Proof.} By multiplying on the left by $A^*$, we have

\be
A^*AB_1+A^*A_1B=0.
\label{eq:tangent5}
\ee
Since $A$ is full rank, $A^*A$ is an invertible $p \times p$ matrix. Thus

\be
B_1=-(A^*A)^{-1}A^*A_1B
\label{eq:tangent6}
\ee
Substituting the above into Equation \ref{eq:tangent3} yields

\be
-A(A^*A)^{-1}A^*A_1B+A_1B=0
\label{eq:tangent7}
\ee
Since $B$ is of full rank, we get

\be
-A(A^*A)^{-1}A^*A_1+A_1=(1-P_A)A_1=0
\label{eq:tangent8}
\ee
which implies that the columns of $A_1$ span the same linear space as the image of $P_A$, i.e. the same space spanned by the columns of $A$. Hence $A_1=AV$ for some $p\times p$ matrix $V$.

\begin{lemma}
The tangent space $T_W(R_p\backslash R_{p-1})$ is spanned by the matrices of the form

\be
AB_1+A_1B
\label{eq:tangent9}
\ee
where $A$ and $B$ are fixed, $AB=W$, and $A_1,B_1$ are $n\times p$ and $p\times n$ matrices, respectively.
\end{lemma}

\noindent
{\bf Proof.}
Define a linear map

\be
\sigma: M^{n\times p}\times M^{p\times n}\to M^{n\times n} \quad {\rm with} \quad  \sigma(A_1,B_1)=AB_1+A_1B
\label{eq:tangent10}
\ee
We have

\be
\dim{\rm Im}(\sigma)=2np-\dim{\rm Ker}(\sigma)
\label{eq:tangent11}
\ee
By the above lemma, $\dim{\rm Ker}(\sigma)=p^2$. Thus the image of $\sigma$ has the same dimension as  the manifold $R_p\backslash R_{p-1}$. Hence all the tangent vectors must be of the form
$AB_1+A_1B$.
\par
\null\par
\noindent
{\bf Corollary.}
The map $\iota$ is of full rank at any point $(A,B)$ where $A$ and $B$ are of rank $p$.

\par
\null\par
\noindent
{\bf Proof.} The space spanned by  all pairs $(A_1,B_1)$ has dimension $2np$. By Lemma 2, the space of all matrices of the form $AB_1+A_1B$ has dimension $2np-p^2$, which is the dimension of $R_p$ by Lemma 1. Since the dimension of the image of the Jacobian of $\iota$ at $(A,B)$ is equal to the dimension of the manifold $R_p$, $\iota$ is of full rank.

We need to prove that $\iota$ is of full rank because we want to measure how far away the function is from being convex. The Hessian of the function $E$ is the sum of two terms, the first of which is positive definite (see Remark 8 below). If $\iota$ were of lesser rank, this first term would contribute less to the total Hessian. In particular, if $\iota$ had rank zero, then the first term of the Hessian would be equal to zero and, as a result, a point where the Hessian is positive would not necessarily exist preventing the existence of a global minimum.

\begin{lemma}
For any $W \in R_p$, there exist an $n\times p$ matrix $A$ and a $p\times n$ matrix $B$ such that $W=AB$. In other words, $\iota$ is a surjective map.
\end{lemma}

\noindent
{\bf Proof.} We use the following singular decomposition of matrices

\be
W=U_1\Lambda U_2,
\ee
where $U_1,U_2$ are unitary matrices and $\Lambda$ is a diagonal matrix. Since $W$ is of rank no more than $p$, we can write $\Lambda$ as

\be
\Lambda=
\begin{pmatrix}
\lambda_1\\&\ddots\\&&\lambda_{p}\\&&&0\\&&&&\ddots\\&&&&&0
\end{pmatrix}
\label{eq:deco1}
\ee
\noindent
Let $(U_1)_p$ represent the first $p$ columns of $U_1$ and let $(U_2)^p$ be the first $p$ rows of $U_2$.  Let $\Lambda_1$ be the first $p\times p$ minor of $\Lambda$. Then

\be
W=(U_1)_p\Lambda_1(U_2)^p
\label{eq:deco2}
\ee
Thus the theorem is proved by letting $A=(U_1)_p\Lambda_1$ and $B=(U_2)^p$.

In general, $R_p$ is {\it not} a manifold. One of the resolution $\tilde R_p$ of $R_p$ is defined as follows

\[
\tilde R_p=\{(A,V)\mid A\in R_p, V\subset ker A^*,\dim V=n-p\}
\]
In this case, $\tilde R_p$ is a manifold and we can extend the function $F$ to $\tilde R_p$ in a natural way: for $(A,V)\in\tilde R_p$, we let $\tilde F_p(A,V)=F_p(A)$.

By the convexity of the quadratic function $\sum ||y_t-Ax_t||^2$, we get the following conclusion

\begin{theorem}
Both $F_p,\tilde F_p$ are convex functions on $R_p\backslash R_{p-1}$. In particular, all critical points of the functions are global minima.
\end{theorem}

\begin{remark}
By the relation $E=F\circ\iota$, we have
\[
D^2E=D^2F(\nabla\iota,\nabla\iota)+\nabla F\circ D^2\iota
\]
The first term on the right-hand side is always nonnegative by the convexity of $F_p$. However the second term can be positive or negative, which partly explains why $E$ is not convex and has many critical points that are saddle points.
\end{remark}

Theorem 11 and 5 are related but one does not imply the other. Theorem 11 shows that for complex-valued autoencoders, the error $E$ has a global minimum. However Theorem 11 does not provide further information about the global minimum, not it implies that that all other critical points are saddle points.
We end this section with the following result.

\begin{theorem}
Let
\[
E(A_1,\cdots,A_k)=\sum ||y_t-A_1\cdots A_k x_t||^2,
\]
where $A_i$ are $(\mu_i,\delta_i)$ matrices. Let
\[
\sigma=\min (\mu_i,\delta_i)
\]
Then
\[
E(A_1,\cdots, A_k)=F_\sigma(A_1\cdots A_k)
\]
\end{theorem}

\noindent
{\bf Proof.} The only non-trivial point is that any rank $\sigma$ matrix can be decomposed into
a product of the form $A_1\cdots A_k$, where $A_j$ is a $\mu_j\times\delta_j$ matrix. For $k=2$, this is just  Lemma 4. For $k>2$, the statement can be proved using mathematical induction.

\section*{Appendix B: Direct Proof of Convergence for the Alternate Minimization Algorithm}

It is expected that starting from any full rank initial matrices $(A_1,B_1)$, if we inductively define

\begin{align*}
& A_{k+1}=\Sigma_{YX}B_k^*(B_k\Sigma_{XX}B_k^*)^{-1}\\
&B_{k+1}=(A^*_{k+1}A_{k+1})^{-1}A^*_{k+1}\Sigma_{YX}\Sigma_{XX}^{-1},
\end{align*}
then $(A_k,B_k)$ should converge to  a critical point of $E$.  In this section, we prove the following

\begin{theorem}
In the auto-associative case, assume that

\be
\sum_{j\in\mathcal I}\lambda_j
\label{eq:}
\ee
 are different for different set $\mathcal I$, where $\mathcal I$ is defined in Theorem~\ref{thm4}. Then $(A_k,B_k)$ converges to a critical point of $E(A,B)$.
\end{theorem}

\begin{remark}
The assumption in Theorem 13 is a technical assumption to separate the distortion values of non equivalent critical point. In fact, using Theorem 4, it is equivalent to assuming that each equivalence class of critical points is associated with a different distortion level which characterizes the corresponding critical points.
\end{remark}

\noindent
{\bf Proof.}
In what follows, we use the Hilbert-Schmidt norm of a matrix:
\[
||A||=\sqrt{Tr(A^*A)}
\]
In the auto-associative case, the algorithm becomes
\begin{align*}
& A_{k+1}=\Sigma B_k^*(B_k\Sigma B_k^*)^{-1}\\
&B_{k+1}=(A^*_{k+1}A_{k+1})^{-1}A_{k+1}^*
\end{align*}
The proof of the theorem is in three steps. First, we prove that the sequences $||A_k||$ and $||B_k||$ are bounded so they both have limiting points; second, we prove that the limiting points must be nonsingular matrices if the initial $A_1$ and $B_1$ are nonsingular; finally, we prove that the sequences $A_k$ and $B_k$ are actually convergent under the assumption of the theorem.\\

\noindent
{\bf Step 1.} Substituting $A_{k+1}$ in the definition of $B_{k+1}$, we obtain

\[
B_{k+1}=(B_k\Sigma B_k^*)(B_k\Sigma^2 B_k^*)^{-1}B_k\Sigma
\]
Substituting the expression of $B_k$ into the definition of $A_{k+1}$,  we obtain
\[
A_{k+1}=\Sigma A_k(A_k^*\Sigma A_k)^{-1}(A_k^*A_k)
\]
Since $\Sigma B_k^*(B_k\Sigma^2 B_k^*)^{-1}B_k\Sigma$ is a projection operator, we have

\be
||B_{k+1}|\leq||B_k|| 
\label{eq:convergence100}
\ee
Similarly, if we write $\Sigma=\Sigma_1^2$ for a positive symmetric matrix $\Sigma_1$, we obtain
\[
\Sigma_1^{-1}A_{k+1}=\Sigma_1A_k(A_k^*\Sigma A_k)^{-1}A_k^*\Sigma^*_1\Sigma_1^{-1}A_k
\]
and hence $||\Sigma_1^{-1}A_{k+1}||\leq||\Sigma_1^{-1}A_k||$. Thus we conclude that both $A_k$ and $B_k$ are bounded sequences.
\\

\noindent
{\bf Step 2.} We have $B_kA_{k+1}=I_p$, where $I_p$ is the $p\times p$ identity  matrix. Thus by continuity any limiting point of $B_k$ or $A_k$ must be non-singular.\\

\noindent
{\bf Step 3.}
To prove that the set of limiting points of $B_k$ contains only one point, we observe that the sequence $E(A_k,B_k)$ is decreasing. If $B$ and $B'$ are two limiting points of the sequence $B_k$, we must have

\[
E(A,B)=E(A',B')
\]
where $A=\Sigma B^*(B\Sigma B^*)^{-1}$ and $A'=B'\Sigma (B')^*((B')\Sigma (B')^*)^{-1}$.
By the assumption of the theorem (or Equation \ref{eq:global6}),
if $B\neq B'$ we must have $E(A,B)\neq E(A',B')$, which yields a contradiction. Thus $B_k$ and hence $A_k$ must be convergent.

Since the limit $(A,B)$ satisfies the equations
\[
A=\Sigma B^*(B\Sigma B^*)^{-1}\quad {\rm and} \quad B=(A^*A)^{-1}A^*
\]
by Theorem~\ref{thm1} and ~\ref{thm2}, $(A,B)$ must be a critical point of $E$.

Finally, note that if $\Sigma$ is singular or close to singular, or if there are critical points with the same distortion levels, then the algorithm above could run into numerical issues.

\par
\null\par
\noindent
{\bf Examples.}
The convergence can be better seen when $p=1$. Let
\be
\Sigma=
\begin{pmatrix}
\lambda_1\\
&\ddots\\
&&\lambda_n
\end{pmatrix}
\ee
with $\lambda_1>\cdots>\lambda_n$.
 If $p=1$, then there is a sequence $c_k$ of real numbers such that

\be
B_{k+1}=c_k B_1\Sigma^k
\label{eq:appb15}
\ee

Let $B_1=(b_1,\cdots,b_n)$ and let $i$ be the smallest index such that $b_i\neq 0$. Then \[ B_k=c_k(\lambda_1^kb_1,\cdots,\lambda_n^k b_n) \]
Since $||B_{k+1}||\leq ||B_k||$ (Equation \ref{eq:convergence100}), the sequence $c_k\lambda_i^k$ is bounded for $k\to\infty$. It follows that for any $j>i$, $c_k\lambda_j^kb_j\to 0$ as $k\to\infty$.
Therefore, using Equation \ref{eq:convergence100} again, we have 
$b_k\to ce_i$ for some constant $c$, with $e_1,\cdots, e_n$ denoting the standard basis of ${\mathbb C}^n$. Moreover, $c=b_i$ by a straightforward computation.

The case of arbitrary $p$ values can be addressed using the above example: let $j<i$ and $i-j+1=p$. Let

\be
B_1=
\begin{pmatrix}
0&\cdots&0&b_i&\cdots & b_n\\
&&&  e_j\\
&&&\vdots\\
&&&  e_{i-1}
\end{pmatrix}
\ee
be a matrix of rank $p$ with the same matrix $\Sigma$ as above. Then

\be
B_k\to
\begin{pmatrix}
 e_i\\
e_j\\
\vdots\\
e_{i-1}
\end{pmatrix}
\ee
In conclusion, for any saddle point, one can construct a sequence that converges to it.

\section*{Acknowledgments}
Work in part supported by grants NSF IIS-0513376, NIH LM010235, and NIH-NLM T15 LM07443 to PB,
and NSF DMS-09-04653 to ZL. We wish to acknowledge Sholeh Forouzan for running the simulation for Figure 3.


\bibliographystyle{apalike}
\bibliographystyle{natbib}
\bibliography{baldi,nn,hmm,math,info,biblio,ilab,bibliob,dec,geneexp,bayes}


\end{document}